\documentclass[10pt,twocolumn,letterpaper]{article}

\usepackage[pagenumbers]{wacv} 

\usepackage{amsmath}
\usepackage{amssymb}
\usepackage{bm}
\usepackage{multirow}
\usepackage{float}

\graphicspath{%
  {./diagrams/results/}%
  {./diagrams/}%
  {Wacv/diagrams/results/}%
  {Wacv/diagrams/}%
}

\newcommand{\R}{\mathbb{R}}
\newcommand{\SET}[1]{\mathcal{#1}}
\newcommand{\vp}{\mathbf{p}}
\newcommand{\vn}{\mathbf{n}}
\newcommand{\vg}{\mathbf{g}}
\newcommand{\vx}{\mathbf{x}}
\newcommand{\vc}{\mathbf{c}}
\newcommand{\norm}[1]{\left\lVert#1\right\rVert}
\newcommand{\abs}[1]{\left\lvert#1\right\rvert}
\newcommand{\lblflat}{\textsc{flat}}
\newcommand{\lbledge}{\textsc{edge}}
\newcommand{\lblcurved}{\textsc{curved}}
\definecolor{wacvblue}{rgb}{0.21,0.49,0.74}
\usepackage[pagebackref,breaklinks,colorlinks,allcolors=wacvblue]{hyperref}

\title{SHIFT: Surface-aware High-speed Integration For TSDFs}

\author{Ayaan Choudhury\\
Indian Institute of Technology Jodhpur\\
{\tt\small b23me1013@iitj.ac.in}
\and
Lokender Tiwari\\
BotLab Dynamics\\
{\tt\small lokender.work@gmail.com}
}

\begin{document}
\maketitle
\begin{abstract}
    Real-time 3D mapping is fundamental for autonomous robotic navigation, with Euclidean Signed Distance Fields (ESDFs) serving as the standard representation for online motion planning. While recent advancements in non-projective distance fields yield highly accurate maps, their computational overhead remains a severe bottleneck. Conventional integrators redundantly re-fuse millions of depth pixels every frame, even long after the corresponding voxels have converged, wasting significant computational resources in environments dominated by large planar surfaces. In this paper, we present SHIFT (\textbf{S}urface-aware \textbf{H}igh-speed \textbf{I}ntegration \textbf{F}or \textbf{T}SDFs), an efficient mapping framework designed to reduce this per-frame update cost. By exploiting structural redundancy directly from $3$D depth geometry, SHIFT compresses flat local regions into weighted super-rays and freezes flat-voxel gradients. A compact ESDF voxel layout further reduces the memory footprint of the remaining wavefront. Extensive evaluations across  various RGB-D and LiDAR sequences show that SHIFT cuts TSDF cost by $1.42$ to $4.07$ times, while holding mesh error within millimeters, and reduces ESDF-layer memory by up to $28\%$.  
    \end{abstract}
\section{Introduction}
\label{sec:intro}

It is a core requirement for autonomous robots like micro aerial vehicles (MAVs) navigating clustered indoor spaces and ground robots operating in unstructured outdoor terrain to build accurate and dense representation of their environment online. Among the many map representations available, Euclidean signed distance fields (ESDFs) have become the most practical choice for real-time motion planning because they can be queried directly for both
distance-to-obstacle and, where available, the underlying free-space gradient in constant time, without an expensive nearest-neighbor search over raw geometry~\cite{oleynikova2017voxblox,han2019fiesta}.
However, building such a map means fusing every incoming depth frame into a persistent volumetric structure at sensor rate, on computational budgets that, for a MAV payload, are a fraction of a desktop GPU. Efficiency is therefore not a secondary concern but a hard constraint on what
representation, and what update strategy, is viable at all.

Two broad families address this problem. The first is explicit voxel-grid TSDF/ESDF fusion~\cite{oleynikova2017voxblox,han2019fiesta,pan2022voxfield} and the second is continuous, often learned, distance
fields~\cite{ortiz2022isdf,wu2023log} (\cref{sec:related}). The latter achieve compelling completion in sparsely observed regions, but their training or inference cost still makes sustained real-time operation on MAV
hardware difficult. Explicit voxel-grid pipelines and Voxfield's
non-projective formulation~\cite{pan2022voxfield} remain
the most practical foundation for onboard, hard-real-time mapping, and are our starting point.

However, this foundation has an overlooked inefficiency. A conventional integrator treats every pixel of every incoming frame identically. It re-projects, re-weighs and re-fuses each measurement into its voxel regardless of whether that voxel's distance and gradient have already
converged. In practice, a large share of any indoor and many outdoor scenes is made up of large, locally planar surfaces like walls, floors, ceilings where once enough coplanar measurements have been fused, they no
longer change. Continuing to re-fuse these pixels on every subsequent frame is waste of compute cost. The TSDF update, the gradient re-estimation, and the
downstream bookkeeping are all repeated for a result that does not change. That cost scales with the number of planar pixels although it is unnecessary.

Prior work on mapping efficiency has largely attacked a different axis of the problem including cheaper data structures, GPU parallelism, and more compact map models (\cref{sec:related:accel}). These approaches make each per-pixel or per-voxel operation cheaper, or run more of them in parallel, but still perform that operation for every measurement on every frame. None exploits the fact that on planar-dominated scenes, much of that work is redundant and can be skipped outright. This makes such gains complementary to reducing the amount of work in the first place.

We close this gap with SHIFT (\textbf{S}urface-aware \textbf{H}igh-speed
\textbf{I}ntegration \textbf{F}or \textbf{T}SDFs), a mapping framework that extends Voxfield's non-projective formulation with a processing path that turns planar structure into an explicit source of computational savings
rather than treating it like every other pixel. SHIFT labels each pixel of an organized depth frame as flat, edge, or curved from its 3D depth geometry alone, using a lightweight classifier that requires no RGB or learned semantics and is shared between RGB-D and spinning-LiDAR inputs.
The flat pixels of a local tile are then compressed into a single weighted super-ray, so that integration cost scales with the number of observed surfaces rather than the number of pixels. During TSDF fusion, a flat voxel
whose accumulated weight has crossed a threshold freezes its gradient and on subsequent frames, skips the weighted average whenever the new distance already agrees within a tolerance. A compact ESDF voxel layout further
reduces the memory footprint of the remaining wavefront. Every mechanism is conservative in the same way, the pixels that fail the planarity tests are fused exactly as in Voxfield, and a measurement that disagrees with the stored
distance by more than the skip tolerance still updates that distance. Across various RGB-D and LiDAR sequences SHIFT reduces TSDF integration cost by $1.42$ to $4.07\times$ over the baseline it extends, holding RGB-D mesh error within $0.19$~cm and improving TSDF RMSE on real LiDAR, with up to $28\%$
less ESDF-layer memory.

\section{Related work}
\label{sec:related}

\subsection{Explicit voxel-grid TSDF/ESDF mapping}
\label{sec:related:explicit}

Voxblox~\cite{oleynikova2017voxblox} popularized the pipeline
for onboard MAV mapping where incoming depth measurements are fused into a
spatially-hashed TSDF voxel grid using a projective distance approximation,
from which an ESDF is derived for planning. FIESTA~\cite{han2019fiesta}
improved the efficiency of this last step with a doubly-linked-list structure and a breadth-first raise/lower wavefront
that updates the ESDF incrementally rather than recomputing it from
scratch. Voxfield~\cite{pan2022voxfield}, which we build directly on,
identified that Voxblox's projective distance approximation degrades
sharply at oblique incidence and replaced it with a non-projective
correction driven by a per-voxel gradient estimate, while retaining
FIESTA-style incremental ESDF maintenance. A number of systems have since
targeted the underlying data structure rather than the fusion model itself. VDBFusion~\cite{vizzo2022sensors} and VDBblox~\cite{bai2023vdbblox} replace
the hashed voxel grid with an OpenVDB backend for faster sparse access. VDB-EDT~\cite{zhu2021vdb} applies the same idea to Euclidean distance
transforms and DB-TSDF~\cite{maese2025db} uses a directional bitmask
encoding to compress the TSDF representation itself. D-LIO~\cite{coto2025d}
shows that a similar TSDF formulation is also effective as the geometric
back-end of a direct LiDAR-inertial odometry system, extending the reach of
this family beyond mapping alone. In these systems, valid
pixels of each frame are typically projected, weighed and fused into their voxel
independently of whether the underlying surface has already converged. Some
integrators merge rays that terminate in the same voxel, but the per-pixel
projection and weighting still runs in full. That is exactly the redundancy
SHIFT targets.

\subsection{Continuous and learned distance fields}
\label{sec:related:continuous}

An alternative line of work replaces the discretized voxel grid with a
continuous function fitted to the observed geometry. Neural
implicit approaches~\cite{ortiz2022isdf, zhong2023shine,jang2024aisdf,yue2025lgsdf} train a
network, hierarchical feature grid or hybrid octree representation online to regress a signed distance directly, while point-based implicit maps ~\cite{pan2024pin, pan2025pings, tian2025miso} couple this idea with large-scale, globally consistent SLAM. HIO-SDF~\cite{vasilopoulos2024hio}
similarly builds a hierarchical, incrementally updated implicit field targeted at online planning. A related family~\cite{wu2023log, le2023accurate, wu2024vdb, ali2024interactive} instead places a Gaussian process prior directly over the distance function. These continuous representations can produce smooth, well-completed fields
even from sparse observations, but their training or inference overhead still makes them difficult to sustain on embedded hardware.

\subsection{Accelerating volumetric mapping}
\label{sec:related:accel}

Several systems accelerate the explicit voxel-grid pipeline itself. nvblox~\cite{millane2024nvblox} re-implements TSDF and ESDF fusion on the
GPU to exploit the parallel nature of per-pixel integration, and a similar GPU-accelerated incremental Euclidean distance transform targets online
motion planning~\cite{9782137}. A unified linear parametric map representation for trajectory planning~\cite{nie2025unified} instead
reduces cost by replacing the dense voxel grid.

\subsection{Plane extraction and structure-aware mapping}
\label{sec:related:planar}

The idea that planar structures dominate man-made indoor and
outdoor scenes has long been exploited for geometric processing. Fast
plane extraction methods for organized point clouds, such as the
agglomerative cell-based clustering of PEAC~\cite{feng2014peac}, segment a
depth image into planar patches in a single pass, and our depth-geometry
surface classifier (\cref{sec:method:classify}) follows the same spirit, working directly on the organized pixel grid and not on unordered point cloud. In the implicit-mapping literature, AiSDF~\cite{jang2024aisdf} similarly uses structural cues for sharper indoor reconstructions, though to improve
completion rather than reduce per-frame cost.

\section{Method}
\label{sec:method}


SHIFT extends Voxfield's non-projective formulation~\cite{pan2022voxfield}
with a processing path that turns planar structure into computational savings.
At frame $k$, the sensor produces an organized point cloud
$\SET{P}_k=\{\vp_{C}(u,v)\}$ on a $W\times H$ pixel grid in the camera frame
$C$, together with the pose $\mathbf{T}_{wk}\in SE(3)$ that maps camera
coordinates into the world frame $w$. RGB-D frames arrive already organized on
the image grid. LiDAR scans are projected into a spherical range image
of the same size using the sensor's vertical field of view
$[\mathrm{fov}_{\mathrm{down}},\,\mathrm{fov}_{\mathrm{up}}]$, so that both
modalities share the same downstream pipeline. The frame is processed in
three stages plus a compact voxel layout: classification into
$\lblflat$/$\lbledge$/$\lblcurved$ segments (\cref{sec:method:classify}),
coplanar bundling of flat pixels into weighted super-rays
(\cref{sec:method:bundle}), and non-projective TSDF fusion with
gradient freezing and update skipping (\cref{sec:method:fuse}),  and a
compact ESDF layout that reduces the wavefront's memory.

We adopt Voxfield~\cite{pan2022voxfield} as the fusion back-end. Two
spatially-hashed voxel maps share a common voxel size $\nu\in\R^{+}$: a TSDF
and the ESDF derived from it. Each TSDF voxel $V_i$ stores a truncated signed
distance $D_i$, an integration weight $W_i$, and a unit gradient (surface
normal) $\vg_i\in\R^{3}$. Incoming measurements are fused with inverse-variance sensor weights. The projective signed distance from a measured point $\vp_j$, sensor origin $\mathbf{s}_k$, to voxel $V_i$ is
\begin{equation}
  \psi_{ijk} = \operatorname{sign}\!\big((\vp_j-\mathbf{s}_k)\cdot(\vp_j-\vx_i)\big)\,\norm{\vp_j-\vx_i}
  \label{eq:psi}
\end{equation}
The incidence-corrected, non-projective distance used by freeze and skip is
\begin{equation}
  d_{ijk}=
  \begin{cases}
    \abs{\cos\theta}\,\psi_{ijk}, & \alpha=0,\\[4pt]
    \big|\tfrac{(\cos\alpha-1)\sin\theta}{\sin\alpha}+\cos\theta\big|\,\psi_{ijk} & \text{else}
  \end{cases}
  \label{eq:nptsdf}
\end{equation}
with $\cos\theta=(\vp_j\cdot\vg_i)/\norm{\vp_j}$ and $\alpha$ the angle between
$\vg_i$ and the per-measurement normal $\vn_j$. Distances are combined by a
truncated weighted average. The ESDF is handled incrementally with a
FIESTA-style raise/lower wavefront~\cite{han2019fiesta}.

\subsection{Depth-geometry surface classification}
\label{sec:method:classify}

Before classification we estimate a per-pixel normal $\vn_j$ with the same finite-difference cross product as Voxfield, evaluated in row-major order with contiguous memory access, which produces bit-identical normals at a faster access pattern on dense RGB-D and partly offsets the front-end cost of classification (\cref{sec:exp:ablation}).

The first stage labels every valid pixel of the organized cloud as $\lblflat$,
$\lbledge$, or $\lblcurved$ from its $3$D depth geometry. Let $z$ denote the
measurement depth at a pixel: camera-frame $z$ for RGB-D, and Euclidean
range $\norm{\vp_C}$ for LiDAR. Sensor noise is modelled as
$\sigma(z)=\max(\sigma_{\min},\,\beta\,z^{2})$, so later tests loosen
automatically for distant, noisier measurements. Occlusion edges are detected over the four image neighbors. For RGB-D, a pixel
is an occlusion edge if its depth differs from a neighbor by more than
\begin{equation}
  \tau_{\mathrm{jump}}(z)=\max\!\big(\tau_0,\,\gamma\,z^{2}\big)
  \label{eq:jump}
\end{equation}
i.e.\ $|\Delta z|>\tau_{\mathrm{jump}}(z)$. For LiDAR, adjacent beams
are tested with Bogoslavskyi's angular connectivity~\cite{7759050}
on the range image. Letting $r_{\mathrm{near}}$ and $r_{\mathrm{far}}$ be the
nearer and farther of two neighboring returns and $\alpha$ the angle between
their rays, the connectivity angle is
\begin{equation}
  \vartheta=\operatorname{atan2}\!\big(r_{\mathrm{far}}\sin\alpha,\,
    r_{\mathrm{near}}-r_{\mathrm{far}}\cos\alpha\big)
  \label{eq:lidar_beta}
\end{equation}
and the pair is marked discontinuous when
$\vartheta<\theta_{\mathrm{lidar}}$. At ranges too small for the angular test,
the jump test of \cref{eq:jump} is used instead.
The image is then partitioned into $P{\times}P$ cells. Each cell accumulates
the first and second moments of its valid, non-edge points,
$\mathbf{S}=\sum_{\vp\in\text{cell}}\vp$,
$\mathbf{Q}=\sum_{\vp\in\text{cell}}\vp\vp^{\top}$, $n_c=\abs{\text{cell}}$,
from which the centroid $\bm{\mu}=\mathbf{S}/n_c$ and covariance
$\mathbf{C}=\mathbf{Q}/n_c-\bm{\mu}\bm{\mu}^{\top}$ are formed. The smallest
eigenpair $(\lambda_0,\mathbf{e}_0)$ of $\mathbf{C}$ is the least-squares plane.
The normal is $\vn=\pm\mathbf{e}_0$ oriented so that $\vn\cdot\bm{\mu}>0$, and
the plane-fit mean squared error is $\mathrm{MSE}=\lambda_0$. A cell is a
\emph{flat} candidate if its edge fraction stays below
$\varepsilon_{\mathrm{edge}}$, it has enough valid non-edge points, and
\begin{equation}
  \lambda_0 \le \big(\kappa_{\mathrm{cell}}\,\sigma(\mu_z)\big)^{2}.
  \label{eq:cellflat}
\end{equation}
Here $\varepsilon_{\mathrm{edge}}$ is zero for dense RGB-D and a small positive
tolerance for sparse LiDAR range images. Flat candidate cells are merged into plane segments by breadth-first
region growing. Moments are additive, so a region's plane is refit exactly from
the merged moments after each accept. A neighbor cell $j$ is merged into
region $R$ if these hold:
\begin{align}
  \text{(normal)}\quad   & \vn_R\cdot\vn_j \ge \cos\theta_{\mathrm{flat}}\\
  \text{(coplanar)}\quad & \abs{\vn_R\cdot(\bm{\mu}_j-\bm{\mu}_R)} \le \kappa_{\mathrm{m}}\,\sigma(\mu_{j,z})\\
  \text{(refit)}\quad    & \lambda_0' \le \big(\kappa_{\mathrm{reg}}\,\sigma(\mu_z')\big)^{2}
  \label{eq:grow}
\end{align}
where $\lambda_0'$ and $\mu_z'$ are the smallest eigenvalue and centroid depth
of the merged moments. The refit test is the guard against curved surfaces because
local cells may look planar, but accumulated curvature inflates the merged MSE
and growth stops. A segment is retained only if it spans at least $N_{\min}$
cells. A crease is marked where two neighboring segments satisfy
$\vn_{s_a}\cdot\vn_{s_b}<\cos\theta_{\mathrm{crease}}$, so distinct planes are
not bundled or frozen across a corner. Per pixel the classifier emits a label
$S(u,v)\in\{\lblflat,\lbledge,\lblcurved\}$ with priority
$\lbledge\!>\!\lblflat\!>\!\lblcurved$, a segment id, and the set of plane
segments $\{(\vn_s,\vc_s,\mathrm{mse}_s)\}$.

\subsection{Coplanar bundling into weighted super-rays}
\label{sec:method:bundle}

\begin{figure}[t]
  \centering
  \includegraphics[width=\columnwidth]{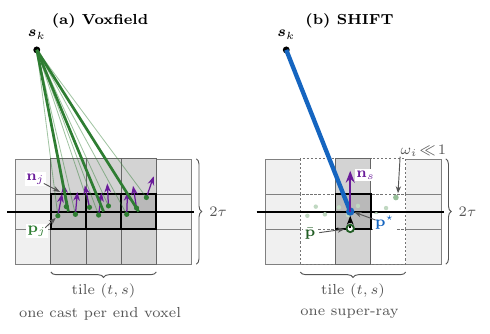}
  \caption{Coplanar bundling on one tile. (a) Each flat pixel is projected, weighed, and cast individually with normal $\vn_j$. (b) The residual-weighted mean $\bar{\vp}$ is projected onto the segment plane (\cref{eq:snap}) and cast once with weight $\omega_{\SET{G}}=\sum_i\omega_i$ and normal $\vn_s$. Shading marks each cast's truncation band $2\tau$. Dashed cells are covered by neighbouring tiles or later frames.}
  \label{fig:superray}
\end{figure}

As shown in \cref{fig:superray}, we replace the many near-identical flat
pixels of a local region by a single representative ray that carries their
combined weight. Before bundling, every flat pixel $i$ belonging to a fitted segment $s$ is
assigned a planarity-residual weight from its point-to-plane distance
$r_i=\abs{\vn_s^{\top}(\vp_i-\vc_s)}$, namely
$\omega_i=(\sigma_{p}/\max(\sigma_{p},\,r_i))^{2}$, where $\sigma_{p}$ is the
residual tolerance. Pixels that fit the plane within $\sigma_{p}$ keep full
weight and the outliers that passed the coarser per-cell / per-region MSE gates are
down-weighted as $\sim 1/r_i^{2}$. Edge and curved pixels keep unit
weight $\omega_i{=}1$.

Flat pixels are grouped by the pair $(t,s)$ of image-tile index $t$ (tiles of
size $P$) and plane-segment id $s$. Grouping by segment as well as tile prevents
a tile straddling two planes from mixing their points into one spurious ray. For
a group $\SET{G}$, we form one super-ray whose point, normal and color
are the $\omega_i$-weighted means of the group, with the mean point then
projected onto the segment's least-squares plane,
\begin{equation}
  \bar{\vp}=\frac{\sum_{i\in\SET{G}}\omega_i\vp_i}{\sum_{i\in\SET{G}}\omega_i}
  \qquad
  \vp^{\star}=\bar{\vp}-\vn_s\big(\vn_s^{\top}(\bar{\vp}-\vc_s)\big)
  \label{eq:snap}
\end{equation}
and whose integration weight is the summed residual weight of the pixels it
represents,
\begin{equation}
  \omega_{\SET{G}}=\sum_{i\in\SET{G}}\omega_i .
  \label{eq:omega}
\end{equation}
Projecting onto the plane in \cref{eq:snap} denoises the depth and gradient and the edge and curved pixels pass through individually at unit weight. A single super-ray thus approximates the group's cumulative contribution with one ray cast.

\subsection{TSDF fusion with freezing and skipping}
\label{sec:method:fuse}

The reduced measurement set is fused with Voxfield's non-projective update,
with each ray's sensor weight scaled by $\omega_{\SET{G}}$ from
\cref{eq:omega} (or $\omega_i$ for an unbundled pixel, including the
planarity-residual weight on flat pixels). Voxel weight $W_i$ is Voxfield's
truncated accumulator, capped at $W_{\max}{=}10^{4}$, and a bundled deposit
therefore matches the pixel-count of the group it represents. The per-ray
label $S$ is written into the voxel as its stored surface type $S_i$. Freeze and skip then apply to every ray,
bundled or not. A voxel is frozen when
\begin{equation}
  \texttt{freeze}(V_i)\;\Longleftrightarrow\;
  S_i{=}\lblflat \,\wedge\, W_i\ge W_{\mathrm{f}} \,\wedge\, \norm{\vg_i}>0 
  \label{eq:freeze}
\end{equation}
and then stops updating its gradient. $W_{\mathrm{f}}{=}5$ is in the same
units as $W_i$, so a well-filled super-ray can cross the threshold in a
single frame, whereas an unbundled ray typically needs several. For a bundled flat voxel
the stored gradient is then the segment normal from the first observing frame
and is not refined. A least-squares plane fit is typically more stable than a
per-voxel gradient, but a first look at grazing incidence is frozen as-is.
There is no residual-motion
unfreeze of the gradient under TSDF fusion (occupancy flips during ESDF sync
do clear the flag). On later frames, the distance $d_{ijk}$ of \cref{eq:nptsdf} is
recomputed with the stored gradient and if
$\abs{d_{ijk}-D_i}<\delta_{\mathrm{skip}}$, the weighted average is skipped.
Otherwise $D_i$ and $W_i$ are still updated, so a surface that moves by more
than the skip tolerance still changes the distance. Finally, we store wavefront indices as $32$-bit
integers rather than $64$-bit, omit unused per-voxel fields and pack
distances, flags and indices first so an active voxel spans fewer cache lines. This does not alter stored distances.

\section{Experiments}
\label{sec:exp}

We evaluate SHIFT on four publicly available datasets
(\cref{tab:datasets}), spanning synthetic and real capture in both modalities. LiDAR scans are projected into a spherical range image so that both modalities share one pipeline. We sweep voxel size and report TSDF integration time and field accuracy against four baselines (\cref{sec:exp:compare}), then a
component ablation (\cref{sec:exp:ablation}) that attributes the measured gain to
the individual stages of \cref{sec:method}. We compare against four established volumetric mapping systems.
\textbf{Voxfield}~\cite{pan2022voxfield} is the primary baseline, sharing
our non-projective formulation and ESDF back-end so any difference isolates
the contribution of the planar processing path. \textbf{Voxblox}~\cite{oleynikova2017voxblox},
\textbf{FIESTA}~\cite{han2019fiesta} and \textbf{VDB-EDT}~\cite{zhu2021vdb}
skip normal estimation entirely, making them cheaper but less accurate, and
locate SHIFT on the same accuracy axis as cheaper projective integrators. We report (i) \emph{TSDF RMSE}, the RMS error of
the fused signed distance at observed voxels versus a ground-truth distance
field (ii) \emph{Mesh$\rightarrow$GT}, the mean distance from extracted mesh
vertices to the nearest ground-truth point (iii) \emph{Chamfer-L1}, the
symmetric mean nearest-neighbour distance between the mesh and the
ground-truth cloud (iv) \emph{reconstruction coverage}, the fraction of
ground-truth points that lie within one voxel of the mesh and
(v) \emph{ESDF-GT RMSE}, the RMS error of the Euclidean field versus
ground-truth distances at observed voxels. 
The primary cost we report is TSDF update time per depth frame, which is the stage SHIFT modifies. ESDF times are per ESDF refresh, not per depth frame as the Euclidean field is rebuilt every five TSDF frames on Flat~\cite{flat} and Newer College~\cite{ramezani2020newer}, every ten on Cow \& Lady~\cite{oleynikova2017voxblox}, and nearly every frame on TUM~\cite{tum}. Sequence ESDF time is therefore \(\mathrm{ms/update}\times N_{\mathrm{ESDF}}\), not \(\times N_{\mathrm{TSDF}}\). End-to-end wall clock is not a
fair headline against the projective baselines, which skip normal estimation entirely, so we compare integration cost directly. We also report resident ESDF-layer memory and process RSS. All methods are implemented in C++ within a common ROS\,2 framework and share the same data loader, pose source, mesh extractor and evaluation code, so that only the mapping back-end differs. The experiments are run on an AMD Ryzen~7 9700X. No GPU is used by any method. Every method is run on a single thread.  SHIFT is run with the classifier cell size of $P{=}12$ for RGB-D and $P{=}8$ for LiDAR. The skip tolerance is $\delta_{\mathrm{skip}}{=}1$~cm, and the freeze threshold is
$W_{\mathrm{f}}{=}5$. Remaining classifier hyperparameters, dataset sequence identifiers, and ground-truth mesh generation are detailed in the supplementary material. Unless noted, quoted per-dataset numbers use a representative voxel size of $10$~cm on RGB-D and $25$~cm on LiDAR. 

\begin{table}[htbp]
\centering
\footnotesize
\setlength{\tabcolsep}{2pt}
\begin{tabular}{lccrrc}
\toprule
Dataset & Sensor & Type & Frames & GT pts & Voxel (cm) \\
\midrule
Flat~\cite{flat}                        & RGB-D & synth. & 399  & 3.20\,M & 5--25 \\
C\&L~\cite{oleynikova2017voxblox}       & RGB-D & real   & 2701 & 0.64\,M & 5--25 \\
TUM~\cite{tum}                          & RGB-D & real   & 84   & 0.66\,M & 5--25 \\
N.~College~\cite{ramezani2020newer}     & LiDAR & real   & 1986 & 2.42\,M & 25--40 \\
\bottomrule
\end{tabular}
\caption{Evaluation datasets. ``Frames'' counts integrated depth frames /
scans. C\&L is Cow \& Lady.}
\label{tab:datasets}
\end{table}

\subsection{Integration cost and accuracy}
\label{sec:exp:compare}

\newlength{\panelw}
\setlength{\panelw}{0.245\textwidth}
\newcommand{\unifimg}[1]{\includegraphics[width=\linewidth]{#1}}

\begin{figure*}[htbp]
\centering
\captionsetup[subfigure]{skip=1pt,font=small,labelformat=empty}
\includegraphics[width=\textwidth]{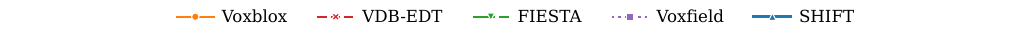}\par\vspace{2pt}
{\small TSDF update (ms/frame) vs.\ voxel size (cm), log $y$; lower is better}\par\vspace{1pt}
\begin{subfigure}[b]{\panelw}\unifimg{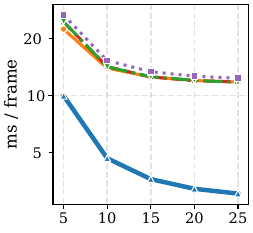}\end{subfigure}\hfill%
\begin{subfigure}[b]{\panelw}\unifimg{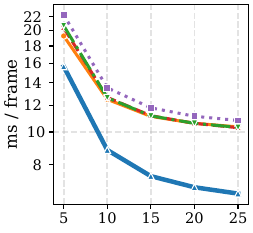}\end{subfigure}\hfill%
\begin{subfigure}[b]{\panelw}\unifimg{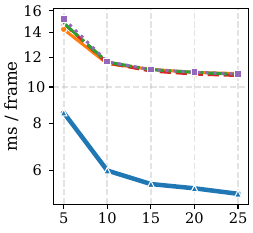}\end{subfigure}\hfill%
\begin{subfigure}[b]{\panelw}\unifimg{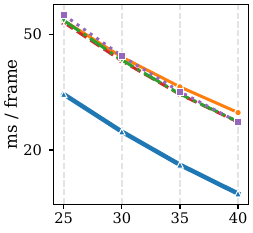}\end{subfigure}

\vspace{3pt}
{\small ESDF update (ms per ESDF refresh) vs.\ voxel size (cm), log $y$; lower is better}\par\vspace{1pt}
\begin{subfigure}[b]{\panelw}\unifimg{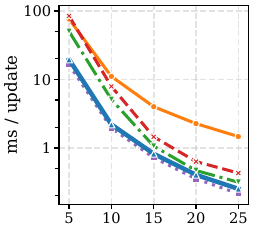}\caption{Flat}\end{subfigure}\hfill%
\begin{subfigure}[b]{\panelw}\unifimg{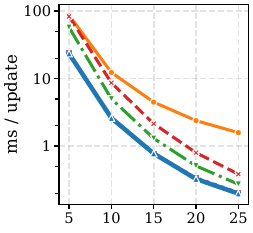}\caption{Cow \& Lady}\end{subfigure}\hfill%
\begin{subfigure}[b]{\panelw}\unifimg{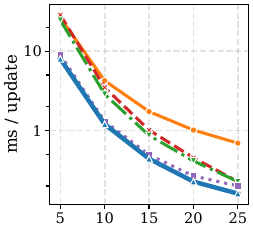}\caption{TUM}\end{subfigure}\hfill%
\begin{subfigure}[b]{\panelw}\unifimg{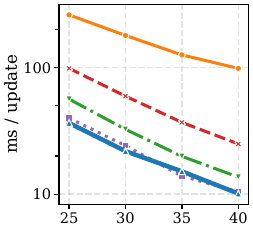}\caption{Newer College}\end{subfigure}
\caption{Integration cost vs.\ voxel size, all methods. Top: TSDF time per depth frame. Bottom: ESDF time per ESDF refresh (not per depth frame). SHIFT (blue) is the lowest TSDF curve on every dataset and every resolution, and its ESDF cost tracks Voxfield while staying well below the projective baselines.}
\label{fig:uniform_runtime}
\end{figure*}

\begin{figure*}[htbp]
\centering
\captionsetup[subfigure]{skip=1pt,font=small,labelformat=empty}
\includegraphics[width=\textwidth]{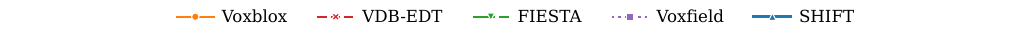}\par\vspace{2pt}
{\small TSDF RMSE (cm) vs.\ voxel size (cm); lower is better}\par\vspace{1pt}
\begin{subfigure}[b]{\panelw}\unifimg{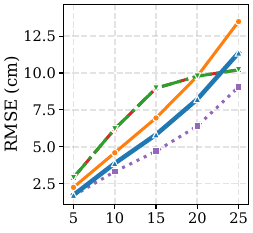}\end{subfigure}\hfill%
\begin{subfigure}[b]{\panelw}\unifimg{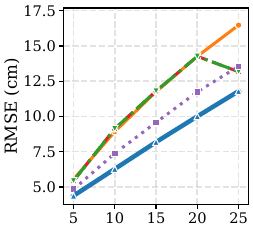}\end{subfigure}\hfill%
\begin{subfigure}[b]{\panelw}\unifimg{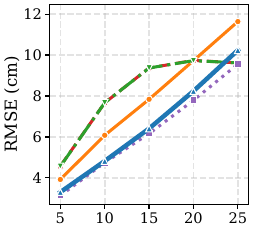}\end{subfigure}\hfill%
\begin{subfigure}[b]{\panelw}\unifimg{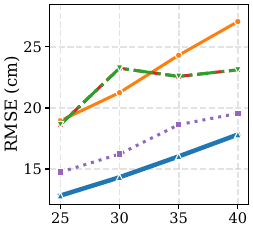}\end{subfigure}

\vspace{3pt}
{\small Mesh$\rightarrow$GT (cm) vs.\ voxel size (cm); lower is better}\par\vspace{1pt}
\begin{subfigure}[b]{\panelw}\unifimg{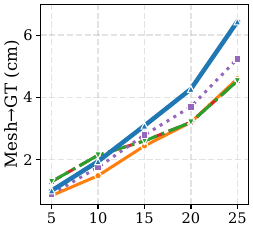}\caption{Flat}\end{subfigure}\hfill%
\begin{subfigure}[b]{\panelw}\unifimg{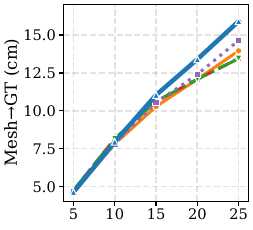}\caption{Cow \& Lady}\end{subfigure}\hfill%
\begin{subfigure}[b]{\panelw}\unifimg{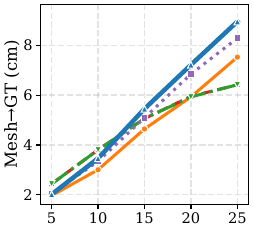}\caption{TUM}\end{subfigure}\hfill%
\begin{subfigure}[b]{\panelw}\unifimg{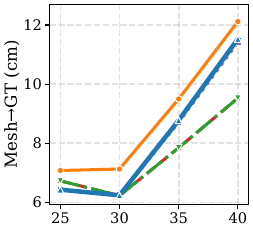}\caption{Newer College}\end{subfigure}
\caption{Accuracy vs.\ voxel size. SHIFT tracks Voxfield on every sequence and stays below the projective
baselines on TSDF RMSE at 10~cm (RGB-D) and 25~cm (LiDAR).}
\label{fig:uniform_accuracy}
\end{figure*}

\begin{figure*}[htbp]
\centering
\includegraphics[width=0.63\textwidth]{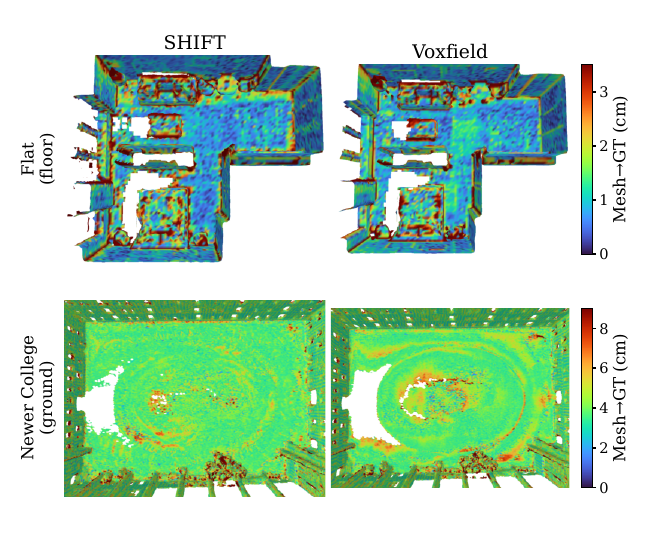}
\caption{Per-vertex Mesh$\rightarrow$GT distance, Ours vs.\ Voxfield, on the
Flat floor @ 10\,cm and the N. College ground plane @ 25\,cm.}
\label{fig:mesherror}
\end{figure*}

\Cref{fig:uniform_runtime} is the main result. SHIFT's TSDF-stage speedup
is the headline, since that is the only stage the method modifies. It is the lowest TSDF curve on all four datasets at every voxel size, cutting TSDF time by $1.53$ to $3.28\times$ at the representative resolution (\cref{tab:sweep}), with the largest gain on the most planar scene and the smallest where planar structure is scarce. ESDF time follows Voxfield throughout and stays well below the projective baselines. Accuracy is essentially unchanged, with a small, bounded cost on synthetic RGB-D (\cref{fig:uniform_accuracy,tab:extraacc,fig:mesherror}). On N. College~\cite{ramezani2020newer}, a real outdoor traverse that includes pedestrians, the two non-projective curves overlap on mesh error ($6.43$~cm) while
SHIFT's TSDF RMSE is lower ($12.82$ vs.\ $14.74$~cm). Projecting each super-ray onto its segment plane denoises real LiDAR. On RGB-D, TSDF RMSE stays below every projective method (Flat $3.87$~cm vs.\ $6.20$~cm for FIESTA / VDB-EDT) and close to Voxfield ($3.87$ vs.\ $3.31$~cm). Mesh error tracks Voxfield on every sequence where the only visible separation is at the coarsest synthetic voxels on Flat, where \cref{eq:snap} costs accuracy.
\Cref{tab:extraacc} reports the other three accuracy metrics at the
representative voxel. ESDF-GT RMSE is within $0.73$~cm of Voxfield on every sequence which is better on TUM and N. College, slightly worse on Flat and C\&L. Coverage drops by at most $0.3$ points (Flat $88.0$ to $87.7$), so fewer rays do not leave a measurable hole in the reconstructed surface. \Cref{fig:mesherror} shows this spatially. Both floors are dominantly blue at well under $1.5$~cm, with warm color confined to wall bases and furniture creases, and the quad ground is near-uniform teal in both panels. The colored quantity is the one-sided Mesh$\rightarrow$GT distance, distinct from the symmetric Chamfer-L1 of \cref{tab:extraacc}, so unmeshed regions (white) reflect
unobserved geometry common to both methods rather than a coverage difference, and the LiDAR scan-pattern rings on N. College are identical between Ours and Voxfield. Crop means differ from the full-mesh values in \cref{tab:ablation}, which exclude unobserved geometry, and we report the latter throughout.

\begin{table}[htbp]
\centering
\footnotesize
\setlength{\tabcolsep}{2pt}
\begin{tabular}{lccccccc}
\toprule
 & & \multicolumn{2}{c}{Chamfer-L1} & \multicolumn{2}{c}{Coverage} & \multicolumn{2}{c}{ESDF-GT} \\
\cmidrule(lr){3-4}\cmidrule(lr){5-6}\cmidrule(lr){7-8}
Dataset & Vox. & Ours & Vf. & Ours & Vf. & Ours & Vf. \\
\midrule
Flat        & 10 & 3.82  & 3.71  & 87.7 & 88.0 & 7.23  & 6.72 \\
Cow \& Lady & 10 & 7.44  & 7.54  & 82.1 & 81.7 & 9.54  & 9.17 \\
TUM         & 10 & 4.74  & 4.71  & 99.7 & 99.7 & 8.24  & 8.37 \\
N.~College  & 25 & 10.37 & 10.39 & 94.7 & 94.6 & 23.35 & 24.08 \\
\bottomrule
\end{tabular}
\caption{Chamfer-L1, reconstruction coverage, and ESDF-GT RMSE at the
representative voxel (``Vox.'', cm). Chamfer-L1 and ESDF-GT in cm, coverage
in \%.}
\label{tab:extraacc}
\end{table}
\Cref{tab:sweep} turns \cref{fig:uniform_runtime} into a ratio against Voxfield. TSDF speedup grows with voxel size on RGB-D, as bundling removes a fixed fraction of rays and the survivors get cheaper as voxels coarsen, but declines on N. College because the LiDAR range image is already coarse relative to the voxel grid at $25$~cm. End-to-end speedup is smaller
throughout, since the classifier is a fixed per-frame cost
(\cref{sec:exp:discussion}).

\begin{table}[htbp]
\centering
\scriptsize
\setlength{\tabcolsep}{0pt}
\resizebox{\columnwidth}{!}{%
\begin{tabular*}{\columnwidth}{@{\extracolsep{\fill}}lrrrrrrr@{}}
\toprule
 & & \multicolumn{2}{c}{TSDF} & \multicolumn{2}{c}{Backend} & \multicolumn{2}{c}{End-to-end} \\
\cmidrule(lr){3-4}\cmidrule(lr){5-6}\cmidrule(lr){7-8}
Dataset & Vox. & Vf. & Ours & Vf. & Ours & Vf. & Ours \\
\midrule
\multirow{5}{*}{Flat}
 & 5  & 26.70 & 9.99\,(2.67$\times$) & 29.95 & 13.87\,(2.16$\times$) & 41.56 & 25.25\,(1.65$\times$) \\
 & 10 & 15.32 & 4.67\,(3.28$\times$) & 15.71 & 5.11\,(3.08$\times$) & 27.38 & 16.55\,(1.66$\times$) \\
 & 15 & 13.38 & 3.62\,(3.70$\times$) & 13.53 & 3.79\,(3.57$\times$) & 25.28 & 15.13\,(1.67$\times$) \\
 & 20 & 12.67 & 3.22\,(3.94$\times$) & 12.74 & 3.30\,(3.86$\times$) & 24.49 & 14.61\,(1.68$\times$) \\
 & 25 & 12.36 & 3.04\,(4.07$\times$) & 12.40 & 3.09\,(4.01$\times$) & 24.07 & 14.51\,(1.66$\times$) \\
\midrule
\multirow{5}{*}{Cow \& Lady}
 & 5  & 22.24 & 15.65\,(1.42$\times$) & 24.64 & 18.03\,(1.37$\times$) & 36.45 & 29.86\,(1.22$\times$) \\
 & 10 & 13.53 & 8.85\,(1.53$\times$) & 13.79 & 9.11\,(1.51$\times$) & 25.74 & 21.10\,(1.22$\times$) \\
 & 15 & 11.80 & 7.41\,(1.59$\times$) & 11.88 & 7.48\,(1.59$\times$) & 23.86 & 19.90\,(1.20$\times$) \\
 & 20 & 11.14 & 6.87\,(1.62$\times$) & 11.18 & 6.90\,(1.62$\times$) & 23.18 & 19.13\,(1.21$\times$) \\
 & 25 & 10.83 & 6.58\,(1.65$\times$) & 10.85 & 6.60\,(1.64$\times$) & 22.81 & 19.37\,(1.18$\times$) \\
\midrule
\multirow{5}{*}{TUM}
 & 5  & 15.21 & 8.54\,(1.78$\times$) & 23.26 & 15.70\,(1.48$\times$) & 35.93 & 29.08\,(1.24$\times$) \\
 & 10 & 11.69 & 5.99\,(1.95$\times$) & 12.86 & 7.06\,(1.82$\times$) & 25.68 & 20.37\,(1.26$\times$) \\
 & 15 & 11.13 & 5.51\,(2.02$\times$) & 11.57 & 5.90\,(1.96$\times$) & 24.38 & 19.40\,(1.26$\times$) \\
 & 20 & 10.96 & 5.37\,(2.04$\times$) & 11.20 & 5.57\,(2.01$\times$) & 23.56 & 19.18\,(1.23$\times$) \\
 & 25 & 10.83 & 5.20\,(2.08$\times$) & 11.01 & 5.33\,(2.07$\times$) & 23.83 & 19.52\,(1.22$\times$) \\
\midrule
\multirow{4}{*}{N.~College}
 & 25 & 58.31 & 31.14\,(1.87$\times$) & 66.29 & 38.43\,(1.72$\times$) & 73.14 & 52.56\,(1.39$\times$) \\
 & 30 & 42.13 & 23.15\,(1.82$\times$) & 46.97 & 27.52\,(1.71$\times$) & 53.83 & 41.63\,(1.29$\times$) \\
 & 35 & 31.75 & 17.81\,(1.78$\times$) & 34.52 & 20.84\,(1.66$\times$) & 41.42 & 34.96\,(1.18$\times$) \\
 & 40 & 24.97 & 14.18\,(1.76$\times$) & 27.06 & 16.20\,(1.67$\times$) & 33.97 & 30.21\,(1.12$\times$) \\
\bottomrule
\end{tabular*} %
}
\caption{Per-frame runtime in ms for Voxfield (``Vf.'') and SHIFT
(``Ours'') across voxel sizes (``Vox.'', cm), with SHIFT's speedup in
parentheses ($>\!1$ means SHIFT is faster). ``TSDF'' is the integration
stage the method modifies; ``Backend'' excludes the shared front end and
end-to-end includes it.}
\label{tab:sweep}
\end{table}

\subsection{Where the saving comes from: ray compression}
\label{sec:exp:rays}

The TSDF integrator is invoked once per surviving ray, so the bundler log is the
right place to measure how much work is removed before fusion. At the
representative voxel size, a Flat frame has $307$\,k input pixels compressed to
$62.8$\,k rays ($4.9\times$), of which $1.7$\,k are super-rays that absorb the
\emph{flat} majority. A N. College scan has $52.9$\,k returns compressed to
$15.9$\,k rays ($3.3\times$), of which $645$ are super-rays
($37.6$\,k \emph{flat} returns, $58$ pixels per super-ray, under the $P{=}8$ tile cap).
A second counter records how many of those rays the merged integrator actually
launches over the full run. Points outside $[\texttt{min},\texttt{max}]$ range
are dropped ($5$~m on RGB-D, $50$~m on LiDAR), then Voxfield's merged caster
keeps one ray per end voxel. Sequence totals are $2.17$\,M vs.\ Voxfield's
$5.31$\,M on Flat ($2.45\times$ fewer) and $22.0$\,M vs.\ $62.0$\,M on N. College ($2.82\times$ fewer). On C\&L, where planar structure is scarce,
those sequence totals are nearly equal ($22.8$\,M vs.\ $25.6$\,M, $1.12\times$).
The bundler ratio exceeds the caster ratio on Flat ($4.9\times$ vs.\ $2.45\times$)
because the merged caster already collapses many coplanar pixels that land in
the same voxel and bundling removes that redundancy in image space first. These compression ratios do not translate one-for-one into the TSDF
speedups of \cref{tab:sweep}. Ray count is not the whole cost as each ray still walks a
truncation band of voxels, and the front end
is untouched. Voxel-update counts versus Voxfield are $1.61\times$ (Flat),
$1.07\times$ (C\&L), $1.10\times$ (TUM) and
$1.83\times$ (N. College). The N. College update-count ratio is close to
the $1.87\times$ TSDF speedup. On RGB-D the TSDF gain is larger than the voxel-update ratio because a
frozen voxel skips gradient re-estimation, so the remaining updates are cheaper.
Bundling removes a large fraction of casts and the remaining per-voxel work is
what the wall clock actually pays. Because the mechanism is driven by coplanar redundancy, the TSDF speedup
tracks how much of each frame the classifier labels \textsc{flat}. $78.9\%$
on Flat, $47.3\%$ on TUM, and $42.1\%$ on C\&L, ranking with the
RGB-D speedups of \cref{tab:sweep}. This ranking breaks on LiDAR, where N. College is $70.6\%$ \emph{flat} but yields only $1.87\times$ because a spherical
range image has fewer pixels to bundle than a dense depth frame. \Cref{tab:memory} quantifies the compact ESDF layout. The layer shrinks by
$8$ to $28\%$ against Voxfield, cutting process RSS by $614$~MB on N. College, though the saving need not reach RSS on the smaller RGB-D maps where
the layer is not the dominant allocation. Flat's RSS is $40$~MB higher, from classifier and bundling buffers Voxfield never allocates.

\begin{table}[htbp]
\centering
\footnotesize
\setlength{\tabcolsep}{3pt}
\begin{tabular}{lrrrrrr}
\toprule
\multirow{2}{*}{Dataset} & \multirow{2}{*}{Vox.} & \multicolumn{3}{c}{ESDF layer (MB)} & \multicolumn{2}{c}{Process RSS (MB)} \\
\cmidrule(lr){3-5}\cmidrule(lr){6-7}
 & & Ours & Voxfield & Ratio & Ours & Voxfield \\
\midrule
Flat        & 10 & 109.7  & 118.8  & 0.92 & 278.0  & 237.6 \\
Cow \& Lady & 10 & 118.8  & 148.5  & 0.80 & 373.4  & 373.1 \\
TUM         & 10 & 119.1  & 148.8  & 0.80 & 283.8  & 280.4 \\
N.~College  & 25 & 1844.3 & 2551.5 & 0.72 & 2227.4 & 2840.9 \\
\bottomrule
\end{tabular}
\caption{Memory footprint at the representative voxel size (``Vox.'', cm).
``Ratio'' is our ESDF layer over Voxfield's.}
\label{tab:memory}
\end{table}

\subsection{Component ablation}
\label{sec:exp:ablation}

To attribute the gain to individual components, we build the method up one
stage at a time on Flat and N. College sequences. From the Voxfield baseline, we add (i) the
classifier without bundling; freeze, skip and residual weights are
already on, but every pixel is still cast as its own ray (ii) super-ray
bundling, the complete SHIFT configuration and (iii) the same configuration with skipping disabled. Results are in \cref{tab:ablation}. Totals are sequence wall clock (front end + TSDF + ESDF). Because ESDF is not refreshed every depth frame, the per-refresh times in \cref{fig:uniform_runtime} must not simply be multiplied by frame count. These runs are independent of the main sweep in \cref{sec:exp:compare}, agreeing to within 2\% on N. College. The no-bundle configuration is not a speedup. On N. College it is slower than
Voxfield (\cref{tab:ablation}). The classifier doubles the front end
($13.6$ to $27.3$~s, ${\sim}6.9$~ms/frame on $53$\,k returns) while TSDF
time is unchanged, because every pixel is still its own ray. Per-pixel normals were already cheap on a sparse scan, so the classifier, range-image
path and angular connectivity test show up in full. On Flat the front end
does not grow ($4.73$ to $4.42$~s over $307$\,k-pixel frames), as the
bit-identical fast normal path of \cref{sec:method:classify} offsets the
classifier on dense RGB-D. The configuration nonetheless changes the fused field
(\cref{tab:ablation}), because
freeze, skip and residual weights act on individual rays. Hence classification is not the source of the
speedup but the enabling stage that makes bundling safe, and the gain comes from bundling itself. Super-rays collapse the TSDF stage $3.28\times$ on Flat
and $1.87\times$ on N. College, turning the front-end deficit into a net end-to-end win in \cref{tab:ablation}. Accuracy is close to Voxfield and improves on LiDAR. N. College TSDF RMSE improves with mesh error unchanged (\cref{tab:ablation}), the residual weighting above accounting for part of the gain and the plane projection of \cref{eq:snap} the rest. On Flat the exchange runs the
other way, trading $0.19$~cm of mesh accuracy for the same $3.28\times$ integration speedup. Skipping is not the source of the speedup costing $1.6\%$ on N. College and nothing on Flat (\cref{tab:ablation}), within accuracy noise. We keep
it as a cheap early-exit on re-observed frozen voxels, not a claimed timing win.

\begin{table}[htbp]
\centering
\footnotesize
\setlength{\tabcolsep}{1pt}
\begin{tabular}{lrrrr}
\toprule
Configuration & Total & TSDF & TSDF & Mesh \\
 & (s) & (s) & RMSE & $\rightarrow$GT \\
\midrule
\multicolumn{5}{l}{\textit{Flat @ 10\,cm (RGB-D)}}\\
Voxfield               & 10.99 & 6.11 & \textbf{3.31} & \textbf{1.75} \\
$+$ Classify only      & 10.87 & 6.30 & 3.48 & 1.90 \\
$+$ Super-rays (SHIFT) & \textbf{6.54} & \textbf{1.89} & 3.87 & 1.94 \\
$-$ Skipping           & 6.55 & 1.89 & 3.88 & 1.94 \\
\midrule
\multicolumn{5}{l}{\textit{N.~College @ 25\,cm (LiDAR)}}\\
Voxfield               & 143.7 & 114.3 & 14.74 & \textbf{6.43} \\
$+$ Classify only      & 153.9 & 112.2 & 13.90 & 6.44 \\
$+$ Super-rays (SHIFT) & \textbf{102.7} & \textbf{60.4} & \textbf{12.82} & \textbf{6.43} \\
$-$ Skipping           & 104.4 & 62.1 & 12.82 & 6.43 \\
\bottomrule
\end{tabular}
\caption{Component ablation on Flat and N.~College. Times in s, errors in cm.
Speedups are not repeated here; see \cref{tab:sweep}. Rows marked $+$ add a
stage cumulatively; $-$~Skipping removes the skip test from full SHIFT and is
not a further stage. Best among the three cumulative configurations in bold.}
\label{tab:ablation}
\end{table}

\subsection{Hyperparameter sensitivity}
\label{sec:exp:sensitivity}

We sweep the
three parameters namely the skip tolerance
$\delta_{\mathrm{skip}}$, the classifier cell size $P$, and the freeze threshold $W_{\mathrm{f}}$ and report the result in \Cref{tab:sensitivity}. Skip and freeze do not move the result in a way we interpret. On both datasets the three skip settings and the two freeze thresholds leave TSDF RMSE at the same centimetre-level value, and total time does not change the ranking against
Voxfield. $W_{\mathrm{f}}{=}5$ vs.\ $8$ is expected to be a no-op as a bundled super-ray deposits the summed pixel weight $\omega_{\SET{G}}$, so both thresholds are typically crossed on the first well-filled flat hit (\cref{sec:method:fuse}). Cell size matters most on LiDAR. Growing $P$ from $8$ to $16$ on N. College costs $8.4$~s and raises TSDF RMSE from $12.82$ to $13.20$~cm, because large cells on a sparse range image over-bundle
the ground plane and merge measurements that are not truly coplanar. On dense RGB-D the same sweep leaves total time and RMSE essentially unchanged.

\begin{table}[htbp]
\centering
\footnotesize
\setlength{\tabcolsep}{1pt}
\begin{tabular}{lrrrrr}
\toprule
& & \multicolumn{2}{c}{Flat @ 10\,cm} & \multicolumn{2}{c}{N.~College @ 25\,cm} \\
\cmidrule(lr){3-4}\cmidrule(lr){5-6}
Param. & Value & Total & RMSE & Total & RMSE \\
\midrule
\multirow{3}{*}{$\delta_{\mathrm{skip}}$ (cm)}
 & 0 & 6.55 & 3.88 & 104.4 & 12.82 \\
 & \textbf{1} & 6.54 & 3.87 & 102.7 & 12.82 \\
 & 3 & 6.54 & 3.82 & 104.1 & 12.83 \\
\midrule
\multirow{3}{*}{Cell $P$}
 & \textbf{8}  & 6.76 & 3.86 & 102.7 & 12.82 \\
 & \textbf{12} & 6.54 & 3.87 & 110.3 & 12.98 \\
 & 16 & 6.72 & 3.81 & 111.1 & 13.20 \\
\midrule
\multirow{2}{*}{Freeze $W_{\mathrm{f}}$}
 & \textbf{5} & 6.54 & 3.87 & 102.7 & 12.82 \\
 & 8 & 6.50 & 3.87 & 102.4 & 12.85 \\
\bottomrule
\end{tabular}
\caption{Hyperparameter sensitivity. Total time in seconds, TSDF RMSE in cm.
Paper defaults are \textbf{bold} in the value column; $P{=}12$ is the RGB-D
default and $P{=}8$ the LiDAR default.}
\label{tab:sensitivity}
\end{table}

\section{Conclusion and Future Work}
\label{sec:exp:discussion}

SHIFT turns planar redundancy into an explicit, $1.42$ to $4.07\times$
source of TSDF savings, at millimeter-level RGB-D accuracy with improved
LiDAR RMSE, and up to $28\%$ less ESDF-layer memory. The ablation
attributes this gain to super-ray bundling rather than to the classifier,
and the result is stable across the sweep of \cref{sec:exp:sensitivity}. Three directions follow from where the method is weakest. The front end is
a fixed per-frame cost bundling cannot shrink, narrowing the gain on less
planar or coarse-voxel scenes. N. College falls to $1.12\times$
end-to-end at $40$~cm despite a $1.76\times$ faster TSDF stage, and
sharing classification with normal estimation could close that gap.
Freezing has no residual-motion unfreeze, so dynamic scenes with an
unfreeze trigger are a natural next step. Finally, \cref{eq:snap}'s plane
projection trades accuracy for speed on low-noise RGB-D, raising Flat mesh
error from $1.75$ to $1.94$~cm, and an adaptive projection could recover it.

\clearpage
\maketitlesupplementary
This supplement provides the classifier hyperparameters, dataset sequence
identifiers, and ground-truth generation procedure referenced in
Section~4 of the main paper, plus additional results. All datasets and metrics follow the main paper. Our code, converted dataset bags, and evaluation scripts will be
released publicly upon publication.

\section{Classifier Hyperparameters}
\label{sec:supp-hparams}

\Cref{tab:supp-hparams} gives the concrete value of every symbol introduced
in Section~3 of the main paper. $P{=}12$ (RGB-D) and $P{=}8$ (LiDAR), $\delta_{\mathrm{skip}}{=}1$\,cm
and $W_{\mathrm{f}}{=}5$ are also reported in Section~4 of the main paper. $\nu$
is voxel size, so truncation distance scales with the swept voxel size
rather than being
fixed. $\varepsilon_{\mathrm{edge}}{=}0$ for RGB-D means no cell is rejected
by the edge-fraction test on dense depth. $\kappa_{\mathrm{cell}}$,
$\kappa_{\mathrm{m}}$, $\kappa_{\mathrm{reg}}$, and $\gamma$ are compiled-in
constants in the classifier implementation, identical across every dataset
and modality.

\begin{table}[htbp]
\centering
\footnotesize
\setlength{\tabcolsep}{3pt}
\begin{tabular}{lcc}
\toprule
Symbol / hyperparameter & RGB-D & LiDAR \\
\midrule
Patch size $P$                                   & 12    & 8 \\
Depth noise factor $\beta$                       & 0.002 & 0.0015 \\
Noise floor $\sigma_{\min}$ (m)                  & 0.004 & 0.004 \\
Edge depth jump $\tau_0$ (m)                     & 0.04  & 0.08 \\
Max.\ edge cell fraction $\varepsilon_{\mathrm{edge}}$ & 0 & 0.5 \\
Connectivity angle $\theta_{\mathrm{lidar}}$ ($^\circ$) & -- & 10 \\
Vertical FOV ($^\circ$, sensor-fixed)            & -- & $[-22.5,22.5]$ \\
Min.\ segment size $N_{\min}$ (cells)            & 6     & 3 \\
Flat-merge angle $\theta_{\mathrm{flat}}$ (rad)  & 0.30  & 0.30 \\
Crease angle $\theta_{\mathrm{crease}}$ (rad)    & 0.45  & 0.45 \\
Planarity tolerance $\sigma_{p}$ (m)             & 0.01  & 0.01 \\
Freeze threshold $W_{\mathrm{f}}$                & 5.0   & 5.0 \\
Skip tolerance $\delta_{\mathrm{skip}}$ (m)      & 0.01  & 0.01 \\
Truncation dist.\ $\tau$ (bundling figure)       & $2\nu$ & $3\nu$ \\
Cell-flat multiplier $\kappa_{\mathrm{cell}}$    & 1.6   & 1.6 \\
Coplanar-merge multiplier $\kappa_{\mathrm{m}}$  & 3.0   & 3.0 \\
Region-refit multiplier $\kappa_{\mathrm{reg}}$  & 1.3   & 1.3 \\
Jump-test quadratic coeff.\ $\gamma$             & 0.006 & 0.006 \\
\bottomrule
\end{tabular}
\caption{Every classifier/fusion symbol from Section~3, mapped to its
launch-time value.}
\label{tab:supp-hparams}
\end{table}

\section{Dataset Sequences}
\label{sec:supp-datasets}

\paragraph{Flat.} A synthetic RGB-D sequence \cite{flat}. 399 depth
frames rendered along a bounded trajectory through a synthetic room with
planar walls, floor, and furniture. Ground truth is the simulator's own
analytic surface sample (10{,}000 points), so no ground-truth generation
step is applied.

\paragraph{Cow \& Lady.} The Cow and Lady RGB-D sequence released with
Voxblox \cite{oleynikova2017voxblox}, played back from its distributed ROS
bag. Ground truth is the laser-scanned reference mesh shipped with that
release and hence no generation step is applied.

\paragraph{TUM.} The fr1/room sequence from the TUM RGB-D SLAM
benchmark \cite{tum} (handheld, real Kinect sensor noise). We convert the
distributed ROS1 bag to ROS2 for playback. The TUM benchmark ships
trajectory ground truth only, not a surface reference, so we accumulate our
own ground-truth cloud as described in \cref{sec:supp-gt}. 84 depth clouds
accumulated in the world frame and voxel-downsampled at $0.02$\,m, yielding
$655$k points.

\paragraph{N.~College.} Here, we use the pre-processed version of the Newer College Dataset~\cite{ramezani2020newer} released by the authors of VDB-GPDF~\cite{wu2024vdb}, played back at half rate. Ground truth is likewise
accumulated as described in \cref{sec:supp-gt}, downsampled at $0.1$\,m,
yielding $2.42$\,M points.

A link to the converted N.~College bag we used, alongside the rest of our
data-preparation and evaluation code, will be included in our public
repository upon publication.

\section{Ground-Truth Generation}
\label{sec:supp-gt}

Flat's ground truth is the simulator's own analytic surface sample, and
Cow \& Lady ships a laser-scanned reference mesh with its release. TUM and the version of N.~College we used do not ship a surface
ground truth. TUM provides only 6-DoF trajectory ground truth, and the version of
N.~College we used has only LiDAR odometry ground truth. So for these two sequences
we accumulate a reference point cloud directly from the sensor data with a
small offline utility.

Given a recorded bag and its trajectory, the utility reads every point
cloud, transforms each into a fixed world frame using the corresponding
pose, and merges the result into a single accumulated cloud. Three
optional controls shape the result: a maximum range that drops points
beyond a chosen distance from the sensor, a stride that keeps only every
$N$-th cloud, and a voxel size that downsamples the accumulated cloud
(disabled entirely if set to zero, keeping the raw accumulation). This script of ground-truth mesh generation will be released upon publication of this work. For
TUM's fr1/room sequence, this produced a reference cloud of 84 clouds
accumulated in the world frame and downsampled at $0.02$\,m, yielding
$655$k points. For N.~College, the same utility produced a reference
cloud downsampled at $0.1$\,m, yielding $2.42$\,M points.

\section{Additional Results}
\label{sec:supp-extra}

\newlength{\suppanelw}
\setlength{\suppanelw}{0.245\textwidth}
\newcommand{\suppimg}[1]{\includegraphics[width=\linewidth]{diagram_supplementary/#1}}

\subsection{Classifier label distribution}
\label{sec:supp-labels}

Section~4 of the main paper reports the flat-pixel fraction per dataset
($78.9\%$ Flat, $47.3\%$ TUM, $42.1\%$ Cow \& Lady, $70.6\%$ N.~College).
\Cref{tab:supp-labels} gives the full three-way split, averaged over every
logged frame at the representative voxel size ($10$\,cm RGB-D, $25$\,cm
LiDAR). These are read directly from the classifier's own per-frame
counters and reported here.

\begin{table}[htbp]
\centering
\begin{tabular}{lrrr}
\toprule
Dataset & \textsc{flat} & \textsc{edge} & \textsc{curved} \\
\midrule
Flat                 & 78.9\% & 4.4\% & 16.7\% \\
Cow \& Lady           & 42.1\% & 2.6\% & 55.3\% \\
TUM                   & 47.3\% & 2.0\% & 50.7\% \\
N.~College  & 70.6\% & 8.8\% & 20.6\% \\
\bottomrule
\end{tabular}
\caption{Per-pixel classifier label distribution, averaged over the logged
frames of the representative-voxel run. LiDAR classifies more edge pixels
than any RGB-D sequence.}
\label{tab:supp-labels}
\end{table}

\subsection{Per-frame timing distribution}
\label{sec:supp-timing}

Section~4 of the main paper reports TSDF update time as a single mean per dataset at the
representative voxel size (Table~3 of main paper). \Cref{tab:supp-timing} gives the full
per-frame distribution behind that mean and standard deviation and
min/max, for the same representative-voxel runs, read from each run's
own per-frame timer. SHIFT's spread is tighter than Voxfield's on Flat and TUM, comparable on
Cow \& Lady. N.~College's per-frame timer was not captured for this launch
configuration, so its distribution is not available.

\begin{table}[htbp]
\centering
\footnotesize
\begin{tabular}{llrrrr}
\toprule
Dataset & Method & Mean & Std & Min & Max \\
\midrule
\multirow{2}{*}{Flat}       & Voxfield & 15.32 & 0.45 & 11.01 & 28.79 \\
                            & SHIFT    & 4.67  & 0.32 & 1.14  & 8.42  \\
\multirow{2}{*}{Cow \& Lady}& Voxfield & 13.53 & 3.39 & 5.35  & 19.32 \\
                            & SHIFT    & 8.85  & 4.19 & 2.41  & 17.19 \\
\multirow{2}{*}{TUM}        & Voxfield & 11.69 & 1.23 & 8.50  & 15.20 \\
                            & SHIFT    & 5.99  & 1.62 & 3.21  & 11.89 \\
\bottomrule
\end{tabular}
\caption{Per-frame TSDF update time (ms) at the representative voxel size
(10\,cm), mean $\pm$ std with [min, max] shown as separate columns.}
\label{tab:supp-timing}
\end{table}

\subsection{Reproducibility}
\label{sec:supp-repro}

All experiments run on Ubuntu 22.04 with ROS~2 Humble, built with C++14
under a single CMake/ament workspace. Beyond the ROS~2 core, the mapping
backend links Eigen3 for linear algebra and Protobuf, gflags, and glog for
configuration and logging. No GPU is used at any point in the pipeline
(main paper, Section~4). Every baseline shares the identical build, data
loader, pose source, mesh extractor, and evaluation code described there,
so the only difference between methods is the mapping back-end itself.

\subsection{Full voxel sweeps of supplementary metrics}
\label{sec:supp-sweeps}

\begin{figure*}[htbp]
\centering
\includegraphics[width=\textwidth]{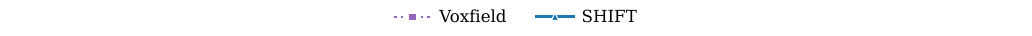}\par\vspace{2pt}
{\small Front-end time (ms/frame) vs.\ voxel size (cm); lower is better}\par\vspace{1pt}
\begin{subfigure}[b]{\suppanelw}\suppimg{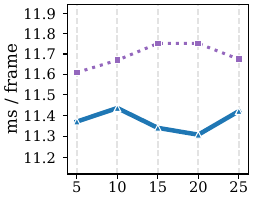}\end{subfigure}\hfill%
\begin{subfigure}[b]{\suppanelw}\suppimg{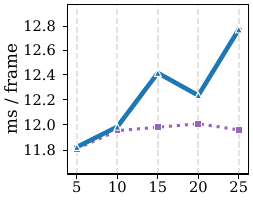}\end{subfigure}\hfill%
\begin{subfigure}[b]{\suppanelw}\suppimg{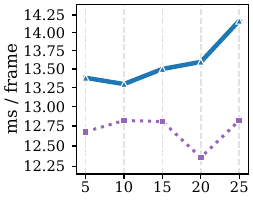}\end{subfigure}\hfill%
\begin{subfigure}[b]{\suppanelw}\suppimg{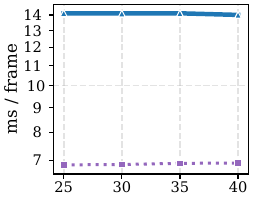}\end{subfigure}

\vspace{3pt}
{\small Chamfer-L1 (cm) vs.\ voxel size (cm); lower is better}\par\vspace{1pt}
\begin{subfigure}[b]{\suppanelw}\suppimg{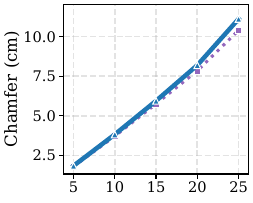}\caption{Flat}\end{subfigure}\hfill%
\begin{subfigure}[b]{\suppanelw}\suppimg{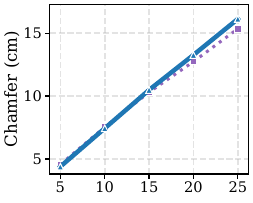}\caption{Cow \& Lady}\end{subfigure}\hfill%
\begin{subfigure}[b]{\suppanelw}\suppimg{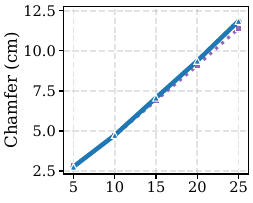}\caption{TUM}\end{subfigure}\hfill%
\begin{subfigure}[b]{\suppanelw}\suppimg{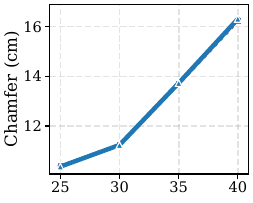}\caption{N.~College}\end{subfigure}
\caption{Front-end time (top) and Chamfer-L1 (bottom) vs.\ voxel size.}
\label{fig:supp-frontend}
\end{figure*}

\begin{figure*}[htbp]
\centering
\includegraphics[width=\textwidth]{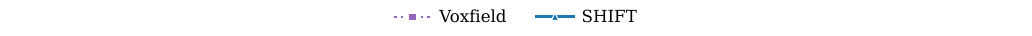}\par\vspace{2pt}
{\small TSDF coverage (\%) vs.\ voxel size (cm); higher is better}\par\vspace{1pt}
\begin{subfigure}[b]{\suppanelw}\suppimg{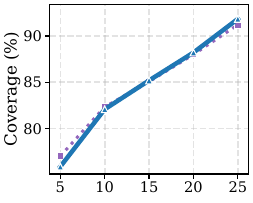}\end{subfigure}\hfill%
\begin{subfigure}[b]{\suppanelw}\suppimg{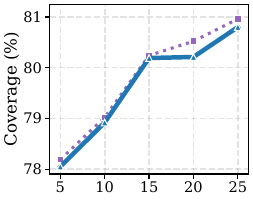}\end{subfigure}\hfill%
\begin{subfigure}[b]{\suppanelw}\suppimg{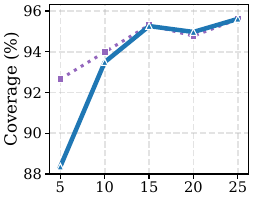}\end{subfigure}\hfill%
\begin{subfigure}[b]{\suppanelw}\suppimg{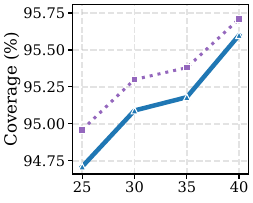}\end{subfigure}

\vspace{3pt}
{\small Reconstruction coverage (\%) vs.\ voxel size (cm); higher is better}\par\vspace{1pt}
\begin{subfigure}[b]{\suppanelw}\suppimg{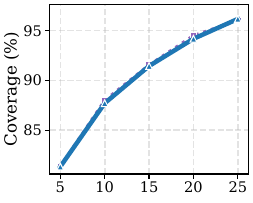}\caption{Flat}\end{subfigure}\hfill%
\begin{subfigure}[b]{\suppanelw}\suppimg{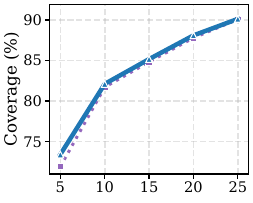}\caption{Cow \& Lady}\end{subfigure}\hfill%
\begin{subfigure}[b]{\suppanelw}\suppimg{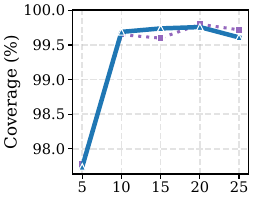}\caption{TUM}\end{subfigure}\hfill%
\begin{subfigure}[b]{\suppanelw}\suppimg{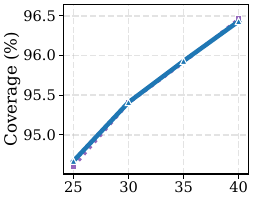}\caption{N.~College}\end{subfigure}
\caption{TSDF coverage (top) and reconstruction coverage (bottom) vs.\ voxel
size.}
\label{fig:supp-coverage}
\end{figure*}

\begin{figure*}[htbp]
\centering
\includegraphics[width=\textwidth]{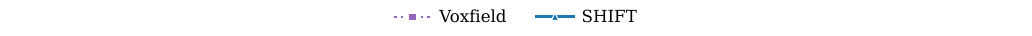}\par\vspace{2pt}
{\small ESDF ground-truth RMSE (cm) vs.\ voxel size (cm); lower is better}\par\vspace{1pt}
\begin{subfigure}[b]{\suppanelw}\suppimg{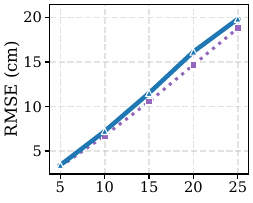}\end{subfigure}\hfill%
\begin{subfigure}[b]{\suppanelw}\suppimg{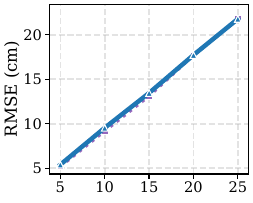}\end{subfigure}\hfill%
\begin{subfigure}[b]{\suppanelw}\suppimg{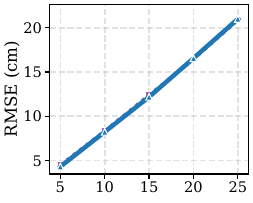}\end{subfigure}\hfill%
\begin{subfigure}[b]{\suppanelw}\suppimg{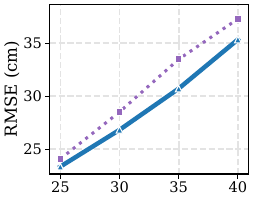}\end{subfigure}

\vspace{3pt}
{\small ESDF-layer memory (MB) vs.\ voxel size (cm); lower is better}\par\vspace{1pt}
\begin{subfigure}[b]{\suppanelw}\suppimg{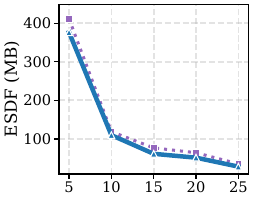}\caption{Flat}\end{subfigure}\hfill%
\begin{subfigure}[b]{\suppanelw}\suppimg{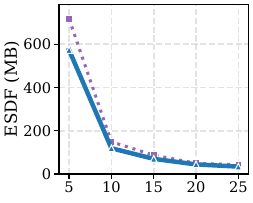}\caption{Cow \& Lady}\end{subfigure}\hfill%
\begin{subfigure}[b]{\suppanelw}\suppimg{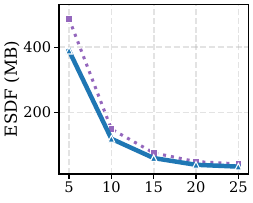}\caption{TUM}\end{subfigure}\hfill%
\begin{subfigure}[b]{\suppanelw}\suppimg{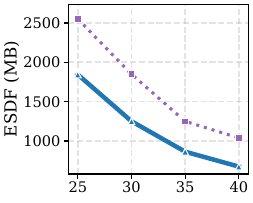}\caption{N.~College}\end{subfigure}
\caption{ESDF ground-truth RMSE (top) and ESDF-layer memory (bottom) vs.\
voxel size. SHIFT is lower than Voxfield at every voxel size on every
dataset for memory.}
\label{fig:supp-esdfgt}
\end{figure*}

\begin{figure*}[htbp]
\centering
\includegraphics[width=\textwidth]{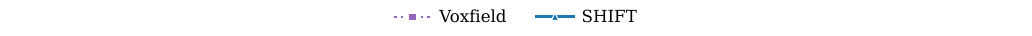}\par\vspace{2pt}
{\small Process RSS (MB) vs.\ voxel size (cm); lower is better}\par\vspace{1pt}
\begin{subfigure}[b]{\suppanelw}\suppimg{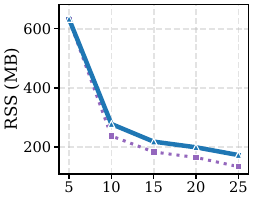}\caption{Flat}\end{subfigure}\hfill%
\begin{subfigure}[b]{\suppanelw}\suppimg{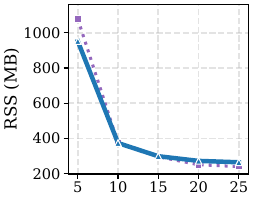}\caption{Cow \& Lady}\end{subfigure}\hfill%
\begin{subfigure}[b]{\suppanelw}\suppimg{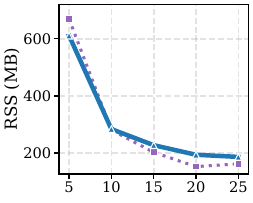}\caption{TUM}\end{subfigure}\hfill%
\begin{subfigure}[b]{\suppanelw}\suppimg{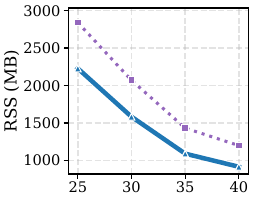}\caption{N.~College}\end{subfigure}
\caption{Process RSS vs.\ voxel size.}
\label{fig:supp-memrss}
\end{figure*}

Section~4 of the main paper reports several metrics per dataset. Here we give the full sweep for the remaining metrics that never appear as a graph
in the main paper: front-end time, Chamfer-L1, coverage, reconstruction
coverage, ESDF ground-truth RMSE, ESDF-layer memory, and process RSS,
across every voxel size and all four datasets
(\cref{fig:supp-frontend,fig:supp-coverage,fig:supp-esdfgt,fig:supp-memrss}).
Of these, ESDF-layer memory (\cref{fig:supp-esdfgt}, bottom) is the only
metric where SHIFT is lower than Voxfield at every voxel size on every
dataset, extending the representative-voxel comparison of Table~4 in the
main paper. Process RSS (\cref{fig:supp-memrss}) and front-end time
(\cref{fig:supp-frontend}, top) are mixed. Voxfield is lower on Flat's
RSS and on front-end time on every sequence but Flat, consistent with
the trade-offs already disclosed in the main paper rather than a new
finding. Chamfer-L1 (\cref{fig:supp-frontend}, bottom), coverage and
reconstruction coverage (\cref{fig:supp-coverage}), and ESDF ground-truth
RMSE (\cref{fig:supp-esdfgt}, top) are near-ties on every dataset,
consistent with the main paper's claim that accuracy is essentially
unchanged.

\subsection{Ablation and sensitivity figures}
\label{sec:supp-ablationfigs}

\begin{figure*}[htbp]
\centering
\begin{subfigure}[b]{0.49\textwidth}
  \includegraphics[width=\linewidth]{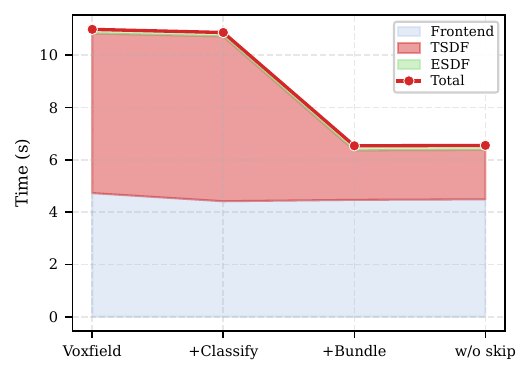}
  \caption{Flat @ 10\,cm}
\end{subfigure}\hfill
\begin{subfigure}[b]{0.49\textwidth}
  \includegraphics[width=\linewidth]{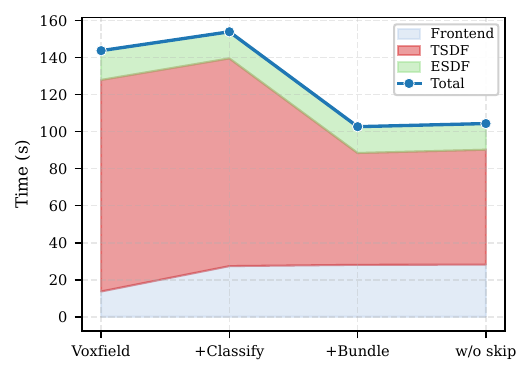}
  \caption{N.~College @ 25\,cm}
\end{subfigure}
\caption{Stage breakdown across the ablation ladder for Flat and N.~College.
Without bundling, the classifier inflates the front end; bundling then
recovers it in the TSDF stage.}
\label{fig:supp-stages}
\end{figure*}

\begin{figure*}[htbp]
\centering
\includegraphics[width=\textwidth]{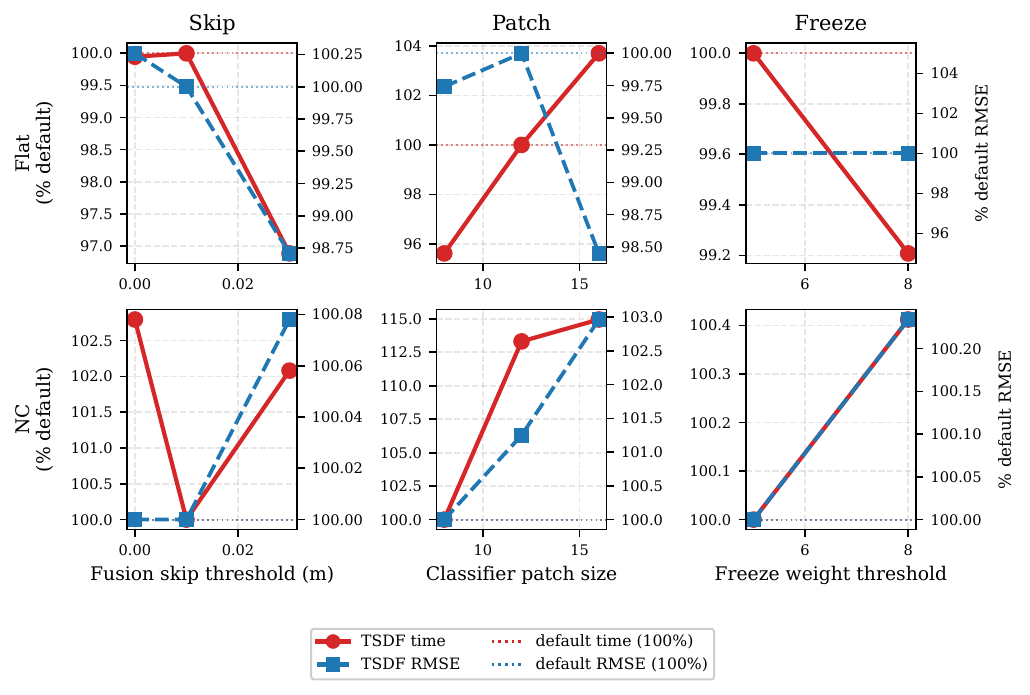}
\caption{Sensitivity of the three main hyperparameters (Table~6
in the main paper), normalized to the default configuration ($=100\%$). Red
is TSDF time, blue is TSDF RMSE. Skip tolerance and freeze threshold are
visually flat; LiDAR classifier cell size is the only knob with a clear
effect.}
\label{fig:supp-sensitivity}
\end{figure*}

The two figures below give a visual counterpart to Table~5 and Table~6 of
the main paper. \Cref{fig:supp-stages} breaks the ablation ladder's total
wall-clock time into its front-end, TSDF, and ESDF shares at each configuration,
showing directly how the classifier's added front-end cost on N.~College
is repaid by the TSDF-stage savings from bundling.
\Cref{fig:supp-sensitivity} renders the main paper's Table~6's numeric hyperparameter sweep
as a normalized curve, making it visually clear that skip tolerance and
freeze threshold barely move the result while the LiDAR classifier cell
size is the one knob with a real effect.

\captionsetup[subfigure]{skip=1pt,font=small,labelformat=empty}

\clearpage
\makeatletter
\setlength{\@colht}{\textheight}
\setlength{\@colroom}{\textheight}
\vsize\textheight
\makeatother
{
    \small
    \bibliographystyle{ieeenat_fullname}
    \bibliography{main}
}

\end{document}